\documentclass{amsart}
\usepackage{amsaddr}
\usepackage{amsmath,mathtools}
\usepackage{amsfonts}
\usepackage[ruled, linesnumbered, noend]{algorithm2e} 
\usepackage{booktabs} 
\usepackage{xcolor}
\usepackage{graphicx}
\usepackage{subfig}
\usepackage{sidecap}
\usepackage{tikz}
\usetikzlibrary{arrows}
\usepackage{vwcol}
\usetikzlibrary{decorations.pathreplacing}
\tikzstyle{arrow} = [thick,->,>=stealth]

\newcommand{\diagentry}[1]{\mathmakebox[1.8em]{#1}}

\title{Winograd Convolution for DNNs: Beyond linear polynomials}
\author{Barbara Barabasz and David Gregg}
\address{School of Computer Science and Statistics,
	Trinity College Dublin,
	Dublin 2, 
	Ireland
}
\email{barabasb@tcd.ie,dgregg@tcd.ie}
\date{}

\begin{document}

\maketitle
\begin{abstract}
    Winograd convolution is widely used in deep neural networks (DNNs). Existing work for DNNs considers only the subset Winograd algorithms that are equivalent to Toom-Cook convolution. We investigate a wider range of Winograd algorithms for DNNs and show that these additional algorithms can significantly improve floating point (FP) accuracy in many cases. 
    We present results for three FP formats: $fp32$, $fp16$ and $bf16$ (a truncated form of $fp32$) using 2000 inputs from the ImageNet dataset. We found that in $fp16$ this approach gives us up to $6.5$ times better image recognition accuracy in one important case while maintaining the same number of elementwise multiplication operations in the innermost loop. 
  In $bf16$ the convolution can be computed using $5\%$ fewer innermost loop multiplications than with currently used Winograd algorithms while keeping the accuracy of image recognition the same as for direct convolution method.
\end{abstract}


\section{Motivation}
In DNNs, and especially in Convolutional Neural Networks (CNNs), a
huge amount of time is spent computing convolution. The simple
\textit{direct} algorithm has a complexity of $O(m^2)$. In contrast,
\textit{fast} convolution algorithms, such as FFT, Toom-Cook, and
Winograd convolution require fewer operations.

The family of Winograd algorithms is based on (1) transforming tiles
of the input and kernel into a modulo polynomials domain (the
\textit{Winograd domain}), where (2) convolution becomes the
\textit{elementwise multiplication} (Hadamard product) with complexity $O(m)$, and (3)
transforming the result back to the orginal domain. Winograd's method
is not a convolution algorithm itself; instead it generates
fast convolution algorithms that operate on fixed-sized tiles of
input, kernel and output.

Winograd's can be used to generate a wide variety of convolution
algorithms with different trade-offs. 
It requires a set of polynomials as input
to generate the convolution algorithm. These polynomials can be linear
($degree=1$) or superlinear ($degree > 1$).

If only linear polynomials are used as inputs, Winograd's method
becomes much simpler, and the resulting algorithms are guaranteed to
need only the theoretically minimum number of operations for the
elementwise multiplication. The set of Winograd algorithms
generated using only linear polynomials is also equivalent to
the set of algorithms that can be generated using the Toom-Cook
method. Toom-Cook is much simpler than the Winograd method,
and as a result it is used to generate the algorithms used
in many implementations of ``Winograd'' convolution.

The selected tile size is critical to the performance of Winograd
convolution. A larger tile size increases the number of
\textit{elementwise multiplication} operations needed for that tile,
but also computes more results per tile. Taking account of extra
operations needed at the boundary of each tile, larger tiles reduce
the number of elementwise multiplication operations per computed
output point. However, the floating point error also grows
exponentially with the tile size \cite{Barabasz18}, so
existing implementations of Winograd for DNNs typically use a small tile size.

In this paper we investigate the effect of higher-order polynomials on
the accuracy of Winograd convolution for DNNs. Our experiments show
that using order-2 polynomials can dramatically reduce the measured
floating point error as compared to linear polynomials. However,
higher order polynomials also increase the required number of
multiplications in Hadamard product. 
This paper addresses the question: Is there a benefit in
using the Winograd method with super-linear polynomials for DNNs,
as compared to the simpler Toom-Cook method? 

We make the following contributions:
\begin{itemize}
\item We demonstrate how the Winograd algorithm with higher-order polynomials can be adapted to DNN convolution
\item We present experimental results for one and two dimensional Winograd convolution, with kernels of the size $3$ (1D) or $3\times 3$ (2D), and find that higher-order polynomials can significantly redude the FP error.
\item We show how using higher-order polynomials offer similar trade-offs between \textit{elementwise multiplications} and FP accuracy, as compared with adjusting the block size.
\item We experimentally identify cases where using higher-order polynomials can improve recognition without increasing \textit{elementwise multiplications} when using half precision ($fp16$), and where  we can improve the performance keeping the accuracy of image recognition in bfloat precision ($bf16$).
\end{itemize}

\section{Toom-Cook versus Winograd algorithm}
\label{sec:T-C_v_Win}
\subsection{Winograd algorithm definition}
Convolution can be expressed as polynomial multiplication. Mapping the elements of kernel vector $h$ and input vector $x$ to coefficients of polynomials $h(a)$ and $x(a)$ respectively, the elements of output vector $s$ (convolution of $h$ and $x$) are equal to the coefficients of polynomial $s(a)=h(a)x(a)$. The Winograd family of algorithms for convolution is based on Chinese Reminder Theorem (CRT) for polynomials \cite{Biggs02}. It says that for polynomial $M(a)$ in ring of polynomials over a field $\mathbb{F}$, $M(a)=m_1(a)\ldots m_\ell(a)$ where $m_i(a)$ are irreducible and pairwise coprime there exists $s(a)$ such as $deg(s(a)) < deg(M(a))$ the unique solution of systems of congruences: $s(a)=s_i(a) \; mod \; m_i(a)$ and  
\begin{equation}
\label{eq:CRT}
s(a)=\sum_i s_i(a)N_i(a)M_i(a) \; mod \; M(a)
\end{equation}
Where $N_i(a)M_i(a)+n_i(a)m_i(a)=1$ and $M_i(a)=M(a)/m_i(a)$.

To compute the result of the convolution - the coefficients of the product of polynomials $h(a)$ and $x(a)$ - we put $s_i(a)=h_i(a)x_i(a) \; mod \; m_i(a)$, $h_i(a)=h(a) \; mod \; m_i(a)$ and  $x_i=x(a) \; mod \; m_i(a)$. 
Operations modulo $m_i(a)$ are equal to finding the remainder from division by $m_i(a)$; so if we assume that all polynomials $m_i(a)$ are of the first degree then the results in modulo $m_i(a)$ arithmetic are all constant polynomials (scalars): $h_i(a)=h(a) \; mod \; m_i(a)=r_h$, $x_i(a)=x(a) \; mod \; m_i(a)=r_x$. 
Then we can perform the computations of $s_i(a)=h_i(a)x_i(a) \; mod \; m_i(a)$ for $i=1,\ldots,\ell$ as single multiplication: $s_i(a)=r_hr_x$. These operations for all $i=1,\ldots,\ell$ are represented by Hadamard product of two vectors consist of elements $h_1(a),\ldots,h_{\ell}(a)$ and $x_1(a),\ldots,x_{\ell}(a)$ (see Figure \ref{fig:T-C}).

The commonly used DNN two-dimensional Winograd convolution algorithm \cite{Lavin16} uses the Matrix Exchange Theorem \cite{Blahut10} and expresses the computations formula for in the following form:
\begin{equation} 
\label{eq:2D_convolution}
A^T(GHG^T \odot B^TXB)A
\end{equation}
Where matrices $H$ and $X$ represents kernel and input values.

In this paper we use the modified version of the Winograd algorithm with polynomial $M(a)$ of degree equal to $deg(s(a))$ and pseudo-point $\infty$.
The exact algorithm to compute matrix elements and a more detailed theoretical description of the method can be found in \cite{Blahut10} \cite{Tolimieri97} \cite{Barabasz18}.

To the best of our knowledge all Winograd algorithms used in DNNs require that all $m_i(a)$ are of the first degree (i.e linear). All such algorithms derived using Winograd's method with linear polynomials can also be found using the Toom-Cook method (\cite{Toom63,Cook66}). Toom-Cook was anlyzed and applied to signal processing problems by S. Winograd in the $1980$s. Winograd also proved that Toom-Cook guarantees that the generated convolution algorithm will use the theoretically minimum possible number of \textit{elementwise multiplications} needed to compute convolution of size $n_o \times n_o$ with a kernel of size $n_h \times n_h$. We denote these algorithms as $F(n_o \times n_o,n_h \times n_h)$) \cite{Winograd80}.

If we use polynomial $m_i(a)$ of degree $d > 1$, then the results of \linebreak $h_i(a)=h(a) \; mod \; m_i(a)$ and $x_i(a)=x(a) \; mod \; m_i(a)$ are polynomials not scalars (see Figure \ref{fig:Winograd}).
Thus to compute $s_i(a)$ we need to multiply two polynomials $h_i(a)$, $x_i(a)$ rather than using simple scalar multiplication. However, to solve this subproblem (i.e. computing the coefficients of the product of two polynomials $h_i(a)$ and $x_i(a)$) we can apply any suitable algorithm, including the Toom-Cook algorithm $F_{T-C}(d\times d,d\times d)$. 
All polynomials $m_i(a)$ used in the Winograd algorithm have to be pairwise coprime, and similarly all polynomials used in the Toom-Cook algorithm to solve the sub-problem also need to be pairwise coprime. But polynomials in the two different groups do not need to be coprime. This means that we can use the same polynomials of the first degree (\textit{points}) in both algorithms. In the Figure \ref{fig:Winograd} we can have $q_i=p_j$. 
Some \textit{points} offer superior floating point accuracy, such as $0$, $-1$ and $1$ (polynomials $a$, $a+1$ and $a-1$) \cite{Barabasz18}.

The approach with polynomials, $m_i(a)$ of degree $d>1$ requires two steps of transformations (see Figure \ref{fig:Winograd}). Firstly, we transform input/kernel into the "Polynomials Winograd domain". That means to transform input/kernel into polynomials of the degree greater than zero. We then transform both those polynomials into scalars in the "Winograd domain". To perform the second transformation we use the Toom-Cook algorithm. Similarly, after computing Hadamard product we first transform the result into "Polynomials Winograd domain" and after this into the original domain. Each of these transforms can be represented by a matrix which multiplied by the input/kernel/output to compute the transformation. We can merge the matrices for these two stages of transformation into a single transformation, allowing us to create three matrices $G^W$, $B^W$ and $A^W$ applied to the kernel, input and result of the Hadamard product respectively. For the clarity, we denote matrices constructed for Toom-Cook algorithm as $G^{(T-C)}$, $A^{(T-C)}$ and $B^{(T-C)}$. 

\begin{SCfigure}[][t]
	\centering
\begin{tikzpicture}[thick,scale=0.7,every node/.style={transform shape}]
\fill[blue!20] (-0.25,0.75) rectangle (0.25,-0.75);
\draw (-0.25,0.75) rectangle (0.25,-0.75);
\draw (-0.25,0.25) -- (0.25,0.25);
\draw (-0.25,-0.25) -- (0.25,-0.25);
\node[draw=none] at (0,1) {$h$};
\draw (2,1.5) rectangle (8,-3); 
\node[draw=none] at (5,1.75) {Winograd domain};
\fill[blue!60] (4,1) rectangle (4.5,0.5);
\fill[blue!60] (4,0) rectangle (4.5,-0.5);
\fill[blue!60] (4,-1) rectangle (4.5,-1.5);
\fill[blue!60] (4,-2) rectangle (4.5,-2.5);
\draw (4,1) rectangle (4.5,0.5);
\draw (4,0) rectangle (4.5,-0.5);
\draw (4,-1) rectangle (4.5,-1.5);
\draw (4,-2) rectangle (4.5,-2.5);
\draw[decorate,decoration={brace,amplitude=5pt}] (0.5,0.75) -- (0.5,-0.75);
\draw[->] (1,0) -- (3.75,0.75) node[above,rotate=25] at (3,0.5) {$a-p_1$};
\draw[->] (1,0) -- (3.75,-0.25) node[above,rotate=-5] at (3,-0.25) {$a-p_2$};
\draw[->] (1,0) -- (3.75,-1.25) node[above,rotate=-25] at (3,-1) {$a-p_3$};
\draw[->] (1,0) -- (3.75,-2.25) node[above,rotate=-35] at (3,-1.75) {$a-p_4$};
\node[draw=none] at (5,0.75) {$\ast$};
\node[draw=none] at (5,-0.25) {$\ast$};
\node[draw=none] at (5,-1.25) {$\ast$};
\node[draw=none] at (5,-2.25) {$\ast$};
\fill[red!60] (5.5,1) rectangle (6,0.5);
\fill[red!60] (5.5,0) rectangle (6,-0.5);
\fill[red!60] (5.5,-1) rectangle (6,-1.5);
\fill[red!60] (5.5,-2) rectangle (6,-2.5);
\draw (5.5,1) rectangle (6,0.5);
\draw (5.5,0) rectangle (6,-0.5);
\draw (5.5,-1) rectangle (6,-1.5);
\draw (5.5,-2) rectangle (6,-2.5);
\fill[red!20] (9.5,1) rectangle (10,-1);
\draw[step=0.5cm] (9.5,1) grid (10,-1);
\draw (9.5,1) -- (9.5,-1);
\node[draw=none] at (9.75,1.25) {$x$};
\draw[decorate,decoration={brace,amplitude=5pt,mirror}] (9.25,1) -- (9.25,-1);
\draw[<-] (6.25,0.75) -- (8.75,0) node[above,rotate=-20] at (7,0.5) {$a-p1$};
\draw[<-] (6.25,-0.25) -- (8.75,0) node[above,rotate=10] at (7,-0.25) {$a-p_2$};
\draw[<-] (6.25,-1.25) -- (8.75,0) node[above,rotate=30] at (7,-1) {$a-p_3$};
\draw[<-] (6.25,-2.25) -- (8.75,0) node[above,rotate=45] at (7,-1.75) {$a-p_4$};
\end{tikzpicture}
\caption{Transformation of the kernel (using matrix $G$) and input (using matrix $B^T$) in one-dimensional Toom-Cook convolution algorithm $F_{T-C}(2\times 2, 3\times3)$ with four root points $p_1$, $p_2$, $p_3$ and $p_4$ (polynomials $a=p_1$, $a-p_2$, $a-p_3$ and $a-p_4$).}
\label{fig:T-C}
\end{SCfigure}
\begin{SCfigure}
	\centering  
\begin{tikzpicture}[thick,scale=0.45,every node/.style={transform shape}]
\fill[blue!20] (-0.25,0.75) rectangle (0.25,-0.75);
\draw (-0.25,0.75) rectangle (0.25,-0.75);
\draw (-0.25,0.25) -- (0.25,0.25);
\draw (-0.25,-0.25) -- (0.25,-0.25);
\node[draw=none] at (0,1) {$h$};
\draw (1.5,2.5) rectangle (13,-4.5);
\node[draw=none] at (8,2.75) {Polynomials' Winograd domain}; 
\draw (4.75,1.5) rectangle (9.75,-4);
\node[draw=none] at (7.75,1.75) {Winograd domain};
\fill[blue!60] (6.25,1) rectangle (6.75,0.5);
\fill[blue!60] (6.25,0) rectangle (6.75,-0.5);
\fill[blue!60] (6.25,-1) rectangle (6.75,-1.5);
\fill[blue!60] (6.25,-2) rectangle (6.75,-2.5);
\fill[blue!60] (6.25,-3) rectangle (6.75,-3.5);
\draw (6.25,1) rectangle (6.75,0.5); 
\draw (6.25,0) rectangle (6.75,-0.5);
\draw (6.25,-1) rectangle (6.75,-1.5);
\draw (6.25,-2) rectangle (6.75,-2.5);
\draw (6.25,-3) rectangle (6.75,-3.5);

\fill[blue!40] (3.25,-1.5) rectangle (3.75,-2.5);
\draw (3.25,-1.5) rectangle(3.75,-2.5); 
\draw (3.25,-2) -- (3.75,-2);
\draw[decorate,decoration={brace,amplitude=5pt}] (0.5,0.75) -- (0.5,-0.75);
\draw[decorate,decoration={brace,amplitude=5pt}] (4,-1.5) -- (4,-2.5);
\draw[->] (0.75,0) -- (6, 0.75) node[above,rotate=10] at (4,0.5) {$a-p_1$};
\draw[->] (0.75,0) -- (6, -0.25) node[above,rotate=-5] at (4,-0.25) {$a-p_2$};
\draw[->] (0.75,0) -- (3,-2) node[above,rotate=-40] at (2.25,-1.25) {$a^2+ba+c$};
\draw[->] (4.25,-2) -- (6,-1.25) node[above,rotate=25] at (5.5,-1.5) {$a-q_1$};
\draw[->] (4.25,-2) -- (6,-2.25) node[above,rotate=-5] at (5.5,-2.25) {$a-q_2$};
\draw[->] (4.25,-2) -- (6,-3.25) node[above,rotate=-35] at (5.5,-3) {$a-q_3$};

\node[draw=none] at (7.25,0.75) {$\ast$};
\node[draw=none] at (7.25,-0.25) {$\ast$};
\node[draw=none] at (7.25,-1.25) {$\ast$};
\node[draw=none] at (7.25,-2.25) {$\ast$};
\node[draw=none] at (7.25,-3.25) {$\ast$};

\fill[red!60] (7.75,1) rectangle (8.25,0.5);
\fill[red!60] (7.75,0) rectangle (8.25,-0.5);
\fill[red!60] (7.75,-1) rectangle (8.25,-1.5);
\fill[red!60] (7.75,-2) rectangle (8.25,-2.5); 
\fill[red!60] (7.75,-3) rectangle (8.25,-3.5);
\draw (7.75,1) rectangle (8.25,0.5); 
\draw (7.75,0) rectangle (8.25,-0.5);
\draw (7.75,-1) rectangle (8.25,-1.5);
\draw (7.75,-2) rectangle (8.25,-2.5);
\draw (7.75,-3) rectangle (8.25,-3.5);

\fill[red!40] (10.75,-1.5) rectangle (11.25,-3);
\draw (10.75,-1.5) rectangle(11.25,-3); 
\draw (10.75,-2) -- (11.25,-2);
\draw (10.75,-2.5) -- (11.25,-2.5);
\draw[decorate,decoration={brace,amplitude=5pt,mirror}] (10.5,-1.5) -- (10.5,-3);
\draw[<-] (8.5,0.75) -- (13.75, 0) node[above,rotate=-10] at (10.75,0.5) {$a-p_1$};
\draw[<-] (8.5,-0.25) -- (13.75, 0) node[above,rotate=5] at (10.75,-0.25) {$a-p_2$};
\draw[<-] (11.5,-2) -- (13.75,0) node[above,rotate=40] at (12.25,-1.25) {$a^2+ba+c$};
\draw[<-] (8.5,-1.25) -- (10.25,-2.25) node[above,rotate=-20] at (9,-1.5) {$a-q_1$};
\draw[<-] (8.5,-2.25) -- (10.25,-2.25) node[above,rotate=5] at (9,-2.25) {$a-q_2$};
\draw[<-] (8.5,-3.25) -- (10.25,-2.25) node[above,rotate=40] at (9,-3) {$a-q_3$};
\node[draw=none] at (14.5,1.25) {$x$};
\fill[red!20] (14.25,1) rectangle (14.75,-1);
\draw (14.25,1) rectangle (14.75,-1);
\draw (14.25,0.5) -- (14.75,0.5);
\draw (14.25,0) -- (14.75,0);
\draw (14.25,-0.5) -- (14.75,-0.5);
\draw[decorate,decoration={brace,amplitude=5pt,mirror}] (14,1) -- (14,-1);
\end{tikzpicture}
\caption{Transformation of kernel (matrix $G$) and input (matrix $B^T$) in $1$ dimensional Winograd convolution algorithm $F_{W}(2\times 2,3\times 3)$ with polynomials $a-p_1$, $a-p_2$, $a^2+ba+c$. The subproblem is solved with Toom-Cook algorithm $F_{T-C}(2\times 2, 2\times2)$ (input $n=3$) and points $q_1$, $q_2$, $q_3$ (polynomials $a-q_1$, $a-q_2$, $a-q_3$).}
\label{fig:Winograd}
\end{SCfigure}

\subsection{Constructing the Transform Matrices}
\subsubsection{Matrices $G^W$ and $A^W$}:
We use the function $\mathrm{vec}(m(a))$ to map the polynomial $m(a)=m_1+m_2a+\ldots+m_na^{n-1}$ to the vector: $\mathrm{vec}(m(a))=\left[\begin{matrix} m_1 & \cdots & m_{n} \end{matrix}\right]^T$
We use $R_{m(a)}[p(a)]$ to denote the remainder from polynomial division of $p(a)$ by $m(a)$.
Rows of the matrix $G^W$ and $A^W$ which stand for transformation with polynomials of the first degree are identical to those in the Toom-Cook algorithm. (Note that we use matrices that are not scaled by factors $N_i$). To construct the submatrices that correspond to the transformation with the polynomial $m_i(a)$ of the degree $d$ higher than one, we have to compose the matrix $G$ with $G'$, where $G'$ represents transformation to the ``Polynomials' Winograd domain'' and the $G$ matrix stands for transformation to the ``Winograd domain'' and is equal to matrix $G^{(T-C)}$ of apropriate size ($F_{T-C}(d\times d,d \times d)$). Analougusly, matrix $A^W=A^{(T-C)}A'$ --- where $A^{(T-C)}$ is generated by the Toom-Cook algorithm $F_{T-C}(d\times d, d\times d)$ --- and $A'$ stands for transformation into the ``Polynomials' Winograd domain'' with polynomial $m_i(a)$ of the degree higher than $1$. The last rows $A_{\ell}$ and $G_{\ell}$ represent the pseudo point $\infty$ needed to construct the modified version of the algorithm (\cite{Blahut10}, \cite{Barabasz18}).
Below we present an example of the construction of matrices $A^W$ and $G^W$ for kernel of size $3\times 3$ and output of size $2\times 2$, choosing polynomials $m_1(a)=a$ and $m_2(a)=a^2+ba+c$. To solve the subproblem $F_{T-C}(2\times 2, 2\times 2)$ we use Toom-Cook algorithm with points $0$, $1$.
\smallskip \\
$
G_2=G^{(T-C)}G'=\left[\begin{matrix}
-1 & 0 \\
1 & 1 \\
0 & 1
\end{matrix}
\right]
\left[\begin{matrix} 
1 & 0 & -c \\
0 & 1 & -b
\end{matrix}
\right]
=
\left[\begin{matrix}
-1 & 0 & c \\
1 & 1 & -b-c \\
0 & 1 & -b
\end{matrix}\right]
$
\smallskip \\
$
A_2=A^{(T-C)}A'=\left[\begin{matrix}
1 & 0 \\
1 & 1 \\
0 & 1
\end{matrix}\right]
\left[\begin{matrix}
1 & 0 & -c & bc \\
0 & 1 & -b & b^2-c
\end{matrix}
\right]
=
\left[\begin{matrix}
1 & 0 & -c & bc \\
1 & 1 & -b-c & bc+b^2-c \\
0 & 1 & -b & b^2-c
\end{matrix}\right] 
$
\smallskip \\
$
G^W=\left[\begin{matrix}
&G_{1}& \\
& G_{2}& \\
0 & 0 & 1
\end{matrix}\right]
=
\left[\begin{matrix}
1 & 0 & 0 \\
-1 & 0 & c \\
1 & 1 & -b-c \\
0 & 1 & -b \\
0 & 0 & 1
\end{matrix}\right] 
$
\smallskip \\
$
A^W=\left[\begin{matrix}
& &A_{1}& \\
& &A_2& \\
0 & 0 & 0 & 1
\end{matrix}\right]
=
\left[\begin{matrix}
1 & 0 & 0 & 0\\
1 & 0 & -c & bc \\
1 & 1 & -b-c & bc+b^2-c \\
0 & 1 & -b & b^2-c \\
0 & 0 & 0 & 1
\end{matrix}\right]  
$
\smallskip \\
The exact algorithms to compute matrices $G^W$ and $B^W$ are presented in algorithm (\ref{alg:Win_G}).
\begin{algorithm}[h]
	\SetAlgoNoLine
	\KwIn{$n_o$ --- size of output, $n_h$ --- size of kernel, $G^{T-C}$, $A^{T-C}$ matrix $G$ and $A$ for $F_{T-C}(d\times d,d \times d)$, $\{m_1(a),\cdots,m_{\ell}(a)\}$ set of $\ell$ irreducible and pairwise coprime polynomials such as $\sum_i deg(m_i(a)) =  n_h+n_o-2$ }
	\KwOut{Matrices $G^W$ and $A^W$ for Winograd convolution}
	$n=n_h+n_o-2$ \\
	\For{$i=1$ to $\ell$}{
		$d =deg(m_i(a))$ \\
    \If{$d==1$}{
    	$p_i=root(m_i(a))$ \\
        $G_{i}=\left[\begin{matrix}
        p_i^0 & \cdots & p_i^{n_h-1}
        \end{matrix}\right]$,  
        $A_{i}=\left[\begin{matrix}
        p_i^0 & \cdots & p_i^{n_o-1}
        \end{matrix}\right]$
        }
    \If{$d>1$}{
    	$G'=\left[\begin{matrix}
    	vec(R_{m_i(a)}[a^0]) & \cdots & vec(R_{m_i(a)}[a^{n_h-1}])
    	\end{matrix}
    	\right]$\\
    	$A'=\left[\begin{matrix}
        vec(R_{m_i(a)}[a^0]) & \cdots & vec(R_{m_i(a)}[a^{n_o-1}])
        \end{matrix}
        \right]$ \\    	
    	$G_{i}=G^{(T-C)}G'$, 
        $A_{i}=A^{(T-C)}A'$
        }
    	$G_{\ell+1} = \left[\begin{matrix}
        0 & \cdots & 0 & 1
        \end{matrix}\right]$, 
        $A_{\ell+1} = \left[\begin{matrix}
        0 & \cdots & 0 & 1
        \end{matrix}\right]$
       	} 
\vspace{5pt}
    $G^W = \left[\begin{matrix}
    G_1\\
    \vdots\\
    G_{\ell}\\
    G_{\ell+1} 
    \end{matrix}\right]$
    \vspace{5pt}
    $A^W = \left[\begin{matrix}
    A_1 \\
    \vdots\\
    A_{\ell} \\
    A_{\ell+1}
    \end{matrix}\right]$
    \caption{Construction of matrix $G^W$ and $A^W$ to transform kernel and result of Hadamerd product in a Winograd convolution algorithm}
    \label{alg:Win_G}
\end{algorithm}

\subsubsection{Matrix $B^W$}
First we construct auxiliary matrix $C$ that includes blocks $C_i$ for $i=1, \cdots, \ell$, where $\ell$ is the number of the polynomials $m_i(a)$. The C matrix represents transformation from the ``Polynomials' Winograd domain'' into the ``Winograd domain''.
The rows stand by transformation with polynomials $m_i(a)$ of the first degree are equal to identity matrix. Blocks stand for transformation with polynomial $m_i(a)$ of degree greater than $1$ represents transformation with matrix $B^{(T-C)}$, generated for subproblem with $F_{T-C}(d \times d, d \times d)$. A second matrix $E$ includes the rest of operations, that is modulo $M(a)$ (remainder) from product of polynomials $M_i(a)$ and the polynomial obtained from extended Euclidean algorithm $N_i(a)$ (see formula \ref{eq:CRT}). Additional zeros in rows of matrix $E$ and column with coefficients of the polynomial $M_i(a)$ implement the modified version of the Winograd algorithm.
\begin{algorithm}[h]
	\SetAlgoNoLine
	\KwIn{$n_o$ - size of output, $n_h$ - size of kernel,  $\{m_1(a),\cdots,m_{\ell(a)}\}$ set of $\ell$ irredicible and pairwise coprime polynomials such as $\sum_i deg(m_i(a)) = n_h+n_o-2$}
	\KwOut{Matrix $B^W$ for Winograd convolution}
	$n=n_h+n_o-2$ \\
	$M(a)=\prod_im_i(a)$ \\
	$M_i(a)=M(a)/m_i(a)$ \\
    \For{$i=1$ to $\ell$}{
    	$d = deg(m_i(a))$ \\
    	\If{$d==1$}{
            $C_{i}=[1]$
    	}
        \If{$d>1$}{
        	$B^{(T-C)}$ matrix $B$ for $F_{T-C}(d \times d,d \times d)$ \\
        	\For{$j=1$ to $2d-1$}{
        		$b_j(a)=B_{1,j}^{(T-C)}+B_{2,j}^{(T-C)}a+ \cdots + B_{2d-1,j}^{(T-C)}a^{2d-2}$
        	}
            $C_{i}=\left[\begin{matrix}
            \mathrm{vec}(R_{m_i(a)}[b_1(a)] & \cdots & \mathrm{vec}(R_{m_i(a)}[b_{2d-1}(a)])
            \end{matrix}\right] 
            $
        }
    }
\vspace{5pt}
    $C=\left[\begin{matrix}
    \diagentry{C_{1}} \\
    &\diagentry{\ddots} \\
    &&\diagentry{C_{\ell}} 
    \end{matrix}\right]$ \\
    \For{$i=1$ to $\ell$}{
    	$N_i(a)$ --- polynomial obtained from extended Euclidean algorithm for polynomials $m_i(a)$ and $M_i(a)$ \\
%
        $E_{i}=\left[\begin{matrix}
        \mathrm{vec}(R_{M(a)}[a^{0}N_i(a)M_i(a)]) \cdots \mathrm{vec}(R_{M(a)}[a^{d-1}N_i(a)M_i(a)])
        \end{matrix}\right]$
    }
\vspace{5pt}
    $E=\left[\begin{matrix}
    E_{1} & \cdots & E_{\ell} \\
    0 & \cdots & 0
    \end{matrix}\right]$ \\
    \vspace{5pt}
    $B^W=\left[\begin{matrix}
      \huge{EC} & vec(M(a))
    \end{matrix}\right]$
    \caption{Construction of matrix B to transform the input in the Winograd convolution algorithm}
    \label{alg:Win_B}
\end{algorithm}

We present an example of constructing matrix $B^W$ for kernels of the size $3\times 3$ and outputs of size $2 \times 2$, chosing polynomials: $m_1(a)=a$ and $m_2(a)=a^2+ba+c$ (as in previous subsection). 
Matrix $B^{(T-C)}$ is generated by the Toom-Cook algorithm $F_{T-C}(2\times 2,2 \times2)$ with points $0,1$.
\smallskip \\
$B^{(T-C)}=\left[\begin{matrix}
-1 & 0 & 0 \\
1 & 1 & -1 \\
0 & 0 &1
\end{matrix}\right]
\qquad 
C_2=
\left[\begin{matrix}
-1 & 0 & -c \\
1 & 1 & -b-1
\end{matrix}\right]
\qquad
C=\left[\begin{matrix}
1 & 0 & 0 & 0  \\
0 & -1 & 0 & -c \\
0 & 1 & 1 & -b-1 
\end{matrix}\right]$
\smallskip \\
Next, we construct the blocks of matrix $E$. The polynomials get from extended Euclidean algorithm \cite{Biggs02} are: $N_1=1$, $N_2=-a$. 
\smallskip \\
$E_1=\left[\begin{matrix}
 1 \\
 b \\
 c 
\end{matrix}\right]
\qquad
E_2=\left[\begin{matrix}
0 & 0\\
0 & c\\
-1 & b
\end{matrix}\right] 
\qquad
E=\left[\begin{matrix}
1 & 0 & 0 \\
b & 0 & c \\
c & -1 & b \\
0 & 0 & 0
\end{matrix}\right] 
$
\smallskip \\
$
EC=\left[\begin{matrix}
1 & 0 & 0 \\
b & 0 & c \\
c & -1 & b \\
0 & 0 & 0
\end{matrix}\right] 
\left[\begin{matrix}
1 & 0 & 0 & 0  \\
0 & -1 & 0 & -c \\
0 & 1 & 1 & -b-1 
\end{matrix}\right]
=
\left[\begin{matrix}
1 & 0 & 0 & 0  \\
b & c & c & -c(b+1)  \\
c & b+1 & b & c-b(b+1)  \\
0 & 0 & 0 & 0 
\end{matrix}\right] 
$
\smallskip \\
$
B^W=\left[\begin{matrix}
1 & 0 & 0 & 0 & 0 \\
b & c & c & -c(b+1) & c \\
c & b+1 & b & c-b(b+1) & b \\
0 & 0 & 0 & 0 & 1
\end{matrix}\right] 
$

\subsection{Optimality of Winograd algorithm}
Toom-Cook algorithms for $2$ dimensional convolution have an optimal number of multiplications $n=(n_h+n_o-1)^2$ for fixed $n_h$ and $n_o$. While computing convolution in DNNs, we break our input into the pieces of the size equal to algorithm input tile. This results in overlap of input tiles at boundaries. The exact number of overlapping input values for whole input depends on the kernel and input/output sizes (see description in \cite{Lavin16}). 
We express the performance of the algorithm as the ratio of the number of multiplications per single output point. Thus, Toom-Cook algorithm $F_{T-C}(2 \times 2,3 \times 3)$ requires $16$ multiplications to compute $4$ output points, so we have ratio equal to $4$. For algorithm $F_{T-C}(4 \times 4, 3 \times 3)$, the $ratio=2.25$. For Toom-Cook convolution with a fixed kernel size the $ratio$ decreases with tile size. The bigger input/output tile, the fewer elementwise multiplications are needed. The elementwise multiplication dominates the execution time of DNN convolution, so reductions in these multiply operations translate to reduced execution time. Unfortunately, with increasing the input/output size the floating point error of the computations increase exponentially \cite{Barabasz18}.

When we apply Toom-Cook algorithm $F_{T-C}(n_o\times n_o,n_h\times n_h)$ the $ratio$ is equal to $(n_h+n_o-1)^2/n_o^2$. In the Winograd method, as we can see from matrix construction, introducing polynomials $m_i(a)$ of the degree greater than $1$ results in larger matrix sizes, which means the bigger number of multiplications. Every Toom-Cook algorithm $F_{T-C}(d\times d,d\times d)$ used to solve subproblem in Winograd algorithm requires $2d-1$ polynomials of the first degree.
The bigger number and higher degree polynomials we use the more multiplications per output point are required.
To compute $F(2\times 2,3\times 3)$ we can use:
\begin{itemize} 
  \item $4$ polynomials of the first degree with $ratio=16/4=4$ (Toom-Cook algorithm)
  \item $2$ polynomials of the first degree and $1$ of the second degree with $ratio=(2+3)^2/4=6.25$ 
  \item $1$ polynomial of the first degree and $1$ of the third degree with $ratio=(1+5)^2/4=9$
  \item $2$ polynomials of the second degree with $ratio=(2*3)^2/4=9$
  \item $1$ polynomial of the fourth degree with $ratio=7^2/4=12.25$ 
\end{itemize}
We can notice that in above example using the polynomial $m_i(a)$ of the $4$th degree do not change input (mapped to the polynomial of the $3$rd degree) and kernel (mapped to the polynomial of the $2$nd degree) pending transformations, so this case only introduce additional multiplications into convolution computations.
Analogously using polynomial $m_i(a)$ of the $3$rd degree does not change the kernel. 
The Winograd method for fixed kernel and output size allows us to construct algorithms with different $ratio$s, while the Toom-Cook method has a constant $ratio$ for given $n_h$ and $n_o$. Thus, for a fixed kernel size, we can construct sets of Winograd matrices with the same $ratio$ but other output/input size. For example for $F_{T-C}(4\times 4,3 \times 3)$ $ratio=36/16=2.25$ and $F_W(6\times 6,3\times 3)$, with $6$ polynomials of the first degree, and one polynomial of the second degree, we have the same $ratio=81/36=2.25$ see table \ref{tab:Winograd_example}. Given these choices with the same computational $ratio$, we can investigate the floating point error of such algorithms and use the more accurate one.	
\begin{table}[t]
	\caption{Number of multiplications for single output point in $2$ dimensional Winograd convolution algorithm for kernel $3\times 3$ and outputs: $2\times 2$, $4\times 4$ and $6\times 6$, for each number of the polynomials of the first and second degree used in CRT. In orange is Toom-Cook algorithm with all polynomials of the first degree.}
	\label{tab:Winograd_example}
\begin{tabular}{|c|ccc|cccc|ccccc|}
\hline
    output size &&$2 \times 2$&&&&$4\times 4$&&&&$6\times 6$&&     \\ 
    \hline  
    No of $m_i(a)$ & \textcolor{orange}{4} & 2 & 0 & \textcolor{orange}{6} & 4 & 2 & 0 & \textcolor{orange}{8} & 6 & 4 & 2 & 0 \\ 
    of degree $1$ &&&&&&&&&&&& \\
    No of $m_i(a)$ & \textcolor{orange}{0} & 1 & 2 & \textcolor{orange}{0} & 1 & 2 & 3 & \textcolor{orange}{0} & 1 & 2 & 3 & 4 \\
    of degree $2$ &&&&&&&&&&&&\\
    Ratio & \textcolor{orange}{4} & 6.25 & 9 & \textcolor{orange}{2.25} & 3.06 & 4 & 5.06 & \textcolor{orange}{1.78} & 2.25 & 2.78 & 3.36 & 4\\
    \hline
\end{tabular}
\end{table}
\section{Tests Results}
\subsection{Random data}
We tested the accuracy of the Winograd convolution algorithm for the kernel of the size $3$ (1D) and $3\times 3$ (2D). We studied a range of output tile sizes from  $2$--$8$ (1D) and $2\times 2$ -- $8\times 8$ (2D). We run our initial experiments over $5000$ loops where kernel and input values were choosen randomly from range $(-1,1)$ with a normal distribution. We computed the Euclidean error of Winograd convolution performed in $fp32$ and compared it with the direct convolution in $fp64$.

\begin{figure}
	\centering 
	\subfloat[]{
		\includegraphics[scale=0.4]{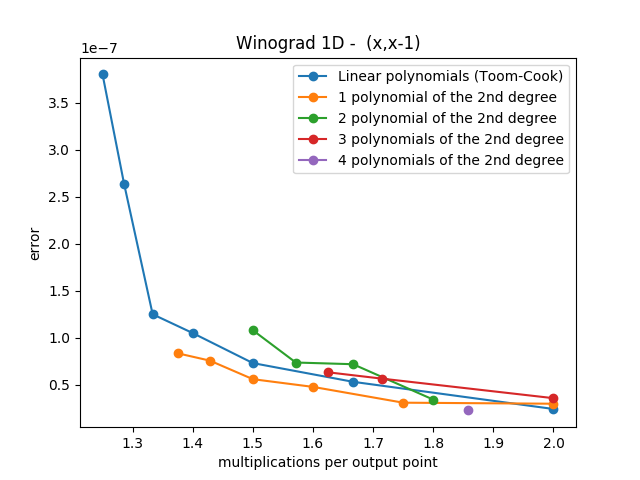}\label{fig:1D}}
	\subfloat[]{
		\includegraphics[scale=0.4]{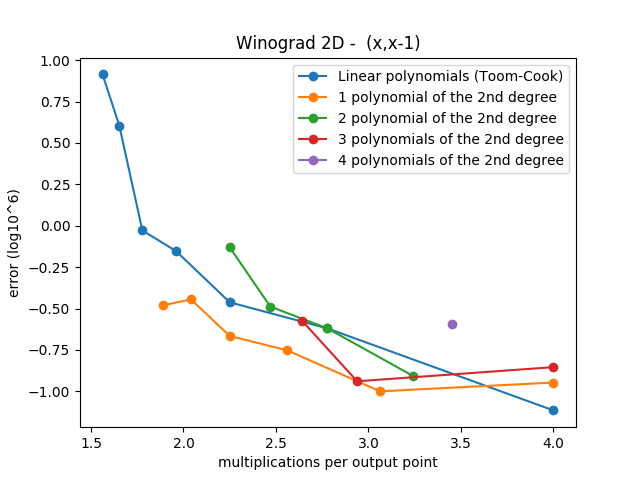}\label{fig:2D}}

\caption{Euclidean error of Winograd convolution in $fp32$ comparing to the direct method computed in $fp64$}
	\label{fig:random}
\end{figure}

We investigated Winograd convolution algorithm with the most promising configurations of polynomials of the first and second degree (as we use the kernel of size $3$ or $3\times 3$). The best results for each computation $ratio$ and polynomial degree configuration are presented in figure \ref{fig:random}. We construct the first degree polynomials using known good root points: $0$, $-1$, $1$, $-1/2$, $2$, $1/2$, $-2$, $-1/4$, $4$ \cite{Barabasz18}. As second degree polynomials, we considered those with the coefficients equal to $0$, $-1$ and $1$, coprime with the polynomials of the first degree. That is: $a^2+1$, $a^2+a+1$ and $a^2-a+1$. To solve the subproblem for the polynomial of degree greater than $1$, we use Toom-Cook convolution algorithm $F_{T-C}(2\times 2, 2\times 2)$ and root points $0$,$-1$ and $\infty$.
In our tests, we noticed that in some cases (up to $ratio$ around $1.9$ for $1D$ and $3.5$ for $2D$) the Winograd algorithm with one polynomial of the second degree gives a smaller floating point error than Toom-Cook, see Figure \ref{fig:random}).
When we use only one polynomial of the second degree we found that $a^2+1$ works the best as it provides only two coefficients, not three.


\subsection{Experiments with real data ImageNet on VGG16}
We next run experiments for the vgg16 CNN \cite{Liu2015} (using Tensorflow Slim) with thirteen 2D convolution layers, with kernel size $3\times 3$. As inputs we use $2000$ images from the ImageNet validation set. The computations were done in $fp32$. We also simulated $fp16$ and $bf16$ by performing the operations in single precision and casting the results to the lower precision.

 We tested the Toom-Cook algorithms with outputs $4\times 4$, $6\times 6$ and $8\times 8$. This means the $ratio$ of multiplications per single ouptut point equals $2.25$, $1.78$ and $1.56$ respectively (see Table \ref{tab:Winograd_example}). For comparision we choose the Winograd algorithm with one polynomial of the second degree for even output sizes, from $6\times 6$ up to $12\times 12$.  The $ratio$ of multiplications per single output point are equal to $2.25$, $1.89$, $1.69$ and $1.56$ respectively.
 In our initial tests on random data, we have found that using the polynomial of the second degree $a^2+1$ works best. Polynomials of the first degree for $F_W(n_o\times n_o,3\times 3)$ were constructed with the root points used for $F_{T-C}((n_o-2)\times (n_o-2),3\times 3)$. 
 For given output and kernel sizes we can construct Winograd algorithms with different computational $ratio$s. 
 In our tests, we used Winograd algorithms with only one polynomial of degree 2. We could achieve better accuracy by using more degree-2, but this would be at the cost of a worse computational $ratio$. We focus on the cases where the image recognition accuracy decreases -- $ratio$ equal to $2.25$ in $fp16$ and $ratio$ between $1.78$ and $1.56$ in $bf16$. We do not present the all possible results, e.g. for $F_W(12\times 12,3\times 3)$ with 2 polynomials of the second degree ($ratio=1.78$), but our results are indicative.
 
 We looked at the percentage of image recognition (top-1) for vgg16 network with Winograd convolution layers in comparison to the same network with direct convolution using the same floating point precision.
 In Tables \ref{tab:recognition_T-C_all}, \ref{tab:recognition_W_all} we present the percentage accuracy of image recognition for different FP precision. For the output sizes we consider, we do not see any changes using $fp32$.
 In $fp16$, all investigated Toom-Cook algorithms failed. In $bf16$ the percentage of image recognition is the same as for direct convolution for Toom-Cook algorithm with output $6\times 6$ ($ratio$ equal to $1.78$), but for output size $8\times 8$ the accuracy decreases.
 
\begin{table}
	\caption{Percentage of image recognition for Toom-Cook convolution algorithm for kernel of the size $3\times 3$ and outputs $4\times 4$, $6\times 6$ and $8\times 8$ in $fp32$, $fp16$ and $bf16$}
	\label{tab:recognition_T-C_all}
	\begin{tabular}{|c|cccc|} 
		\hline 
		method& dir & T-C($4 \times 4$) & T-C($6 \times 6$) & T-C($8 \times 8$) \\
		\hline
		ratio &  & $2.25$ & $1.78$ & $1.56$ \\ 
		$fp32$ & 70 & 70 & 70 & 70 \\
		$fp16$ & 70 & 10 & 0.05 & 0.05\\
		$bf16$ & 70 & 70 & 70 & 68 \\
        \hline
	\end{tabular}
\end{table}
\begin{table}
	\caption{Percentage of image recognition for Winograd convolution algorithm with one polynomial of the second degree $a^2+1$ for kernel of the size $3\times 3$ and outputs $6\times 6$, $8\times 8$, $10\times 10$ and $12\times 12$ in $fp32$, $fp16$ and $bf16$.}
	\label{tab:recognition_W_all}
	\begin{tabular}{|c|ccccc|}
		\hline  
		method & dir & W($6 \times 6$) & W($8\times 8$) & W($10\times 10$) & W($12\times 12$)\\
		\hline
		ratio & &  $2.25$ & $1.89$ & $1.69$ & $1.56$\\
		$fp32$ & 70 & 70 & 70 & 70 & 70 \\
        $fp16$ & 70 & 65 & 0.1 & 0.05 & 0.05 \\
		$bf16$ & 70 & 70 & 70 & 70 & 62 \\
       \hline 
	\end{tabular}
\end{table}
With $fp16$, we see that using Winograd convolution instead of Toom-Cook with the same performance $ratio$ (equal to $2.25$), increases the recognition accuracy from $10\%$ to $65\%$. The main problem we face with $fp16$ is that it cannot store the same range of values as $fp32$. Then using the same good root points (like $0$, $-1$ and $1$) more than once results in lower intermediate values, and less likelihood of overflow.

Using $bf16$, the decrease in image recognition appears for bigger input sizes than in $fp16$. The $bf16$ format allow us to represent nearly the same range of values as single precision. However, the lower number of bits results in lower accuracy of values representation and larger floating point error from operations. In our tests we can observe the impact of this for network with Toom-Cook convolution algorithm with output of the size $8\times 8$. We have not found a configuration of polynomials that would give us the accuracy of image recognition better than $68\%$ with the $ratio$ equal to $1.56$. We construct the Winograd algorithm with the accuracy of image recognition equal to $70\%$ (the same accuracy we get using of a direct convolution algorithm) with $ratio=1.69$. 

\section{Related work}
In the DNN research literature the term ``Winograd convolution algorithm'' is used for both Winograd and Toom-Cook convolution, and in practice the Toom-Cook algorithm is used to generate the convolution matrices.
The general Winograd algorithm described in this paper is not explored a lot in literature. We can find a description of the approach in Winograd \cite{Winograd80}, but not for the multi-channel multiple kernel convolution used for DNNs. A simple example how to construct matrices is presented in \cite{Blahut10}. A more general and detailed description can be found in \cite{Tolimieri97}. Selesnick and Burrus \cite{Burrus94} considered cyclic convolution methods using cyclotomic polynomials in their theoretical work. 

Meng and Brothers in \cite{Meng18} apply the idea of using complex points $i$ and $-i$ (root points of polynomial $a^2+1$) for quantization network. We present a general definition of the method and present floating point accuracy for a couple of different versions of the algorithm. 

There is some work done on the improvement of the FP accuracy of Winograd (Toom-Cook) convolution for DNNs.
Vincent at al. \cite{Vincent17} present the result for one set of matrices, that scaling matrices $G$ and $A^T$ give the more accurate results. Scaling improves the conditioning of used matrices but it is not necessarily always equivalent to decrease the floating point error of computation, particularly for the small size of matrices used in DNNs.    

There are also a couple of methods that allow to increasing the accuracy of dot product computations for matrices transformation, such as more accurate summation algorithms, Strassen matrix multiplication \cite{Zhao18}, etc. However, they require more operations for the transforms, for sorting elements, or for compensated summation, and/or make the implementation more complicated.
In contrast, our approach does not require additional operations for the transformations. 
All of those methods for improving FP accuracy could also be used together with the presented method to reduce floating point error even more. These include pairwise summation over channels, Huffman based summation method and mixed precision computations proposed in \cite{Barabasz18}.

\section{Conclusions}

This paper asks the question: Is there a benefit in using the
Winograd method with superlinear polynomials for DNNs, as compared to
the simpler Toom-Cook method (which is equivalent to Winograd with
linear polynomials)? We describe the construction of Winograd
transformation matrices in general case. We show that the main benefit
of using superlinear polynomials is that the same good root points can be used multiple times, which improves FP
accuracy. The Toom-Cook method allows a trade-off of
elementwise multiplications against FP accuracy by varying
the tile size. The presented Winograd method offers an larger space of
trade-offs between computation and accuracy using higher order polynomials.
Thus, it allows us
find attractive trade-offs that are not available using Toom-Cook.

We find that in $bf16$ precision we can construct an algorithm that
maintains the same accuracy of image recognition as Toom-Cook but has
better $ratio$ of elementwise multiplications per single
output point than Toom-Cook. In $fp16$ precision we can obtain better
accuracy using Winograd convolution algorithm with one polynomial of
the second degree, as compared to Toom-Cook (for the case kernel
$3\times 3$, output $4\times 4$) with the same $ratio$ of number of
elementwise multiplications per output point. The presented
Winograd convolution algorithm does not require additional operations
in the transformation to/from the "Winograd domain", and although the Winograd
method itself is complex, the generated convolution algorithm does not
require a more advanced implementation.

\section*{Acknowledgements}
This work was supported by Science Foundation Ireland grant 12/IA/1381. We also extend our thanks to Andrew Mundy from Arm Research for his contribution.

\bibliographystyle{splncs04}
\bibliography{references}

\begin{thebibliography}{10}
\providecommand{\url}[1]{\texttt{#1}}
\providecommand{\urlprefix}{URL }
\providecommand{\doi}[1]{https://doi.org/#1}

\bibitem{Barabasz18}
Barabasz, B., Anderson, A., Soodhalter, K.M., Gregg, D.: Error analysis and
  improving the accuracy of winograd convolution for dnns. CoRR
  \textbf{abs/1803.10986} (2018), \url{http://arxiv.org/abs/1803.10986}

\bibitem{Biggs02}
Biggs, N.L.: Discrete Mathematics. Oxford University Press, New York, NY, USA,
  2nd. edn. (2002)

\bibitem{Blahut10}
Blahut, R.E.: Fast Algorithms for Signal Processing. Cambridge University
  Press, New York, NY, USA (2010)

\bibitem{Cook66}
Cook, S.A.: On the Minimum Computation Time of Functions. Ph.D. thesis, Harvard
  University, Cambridge, Mass. (1966)

\bibitem{Lavin16}
Lavin, A., Gray, S.: Fast algorithms for convolutional neural networks. In:
  2016 {IEEE} Conference on Computer Vision and Pattern Recognition ({CVPR}).
  pp. 4013--4021. IEEE, Las Vegas, Nevada (2016)

\bibitem{Liu2015}
{Liu}, S., {Deng}, W.: Very deep convolutional neural network based image
  classification using small training sample size. In: 2015 3rd IAPR Asian
  Conference on Pattern Recognition (ACPR). pp. 730--734 (Nov 2015).
  \doi{10.1109/ACPR.2015.7486599}

\bibitem{Meng18}
Meng, L., Brothers, J.: Efficient winograd convolution via integer arithmetic.
  CoRR  \textbf{abs/1901.01965} (2019)

\bibitem{Burrus94}
Selesnick, I.W., Burrus, C.S.: Extending winograd's small convolution algorithm
  to longer lengths. In: 1994 {IEEE} International Symposium on Circuits and
  Systems, {ISCAS} 1994, London, England, UK, May 30 - June 2, 1994. pp.
  449--452 (1994)

\bibitem{Tolimieri97}
Tolimieri, R., An, M., Lu, C.: Algorithms For Discrete Fourier Transform and
  Convolution. Springer-Verlag, New York, NY, USA, 2nd. edn. (1997)

\bibitem{Toom63}
Toom, A.L.: The complexity of a scheme of functional elements realizing
  multiplication of integers. Soviet Mathematics -- Doklady  \textbf{3},
  714--716 (1963)

\bibitem{Vincent17}
Vincent, K., Stephano, K., Frumkin, M., Ginsburg, B., Demouth, J.: On improving
  the numerical stability of winograd convolutions. In: Proceedings of the 5th
  International Conference on Learning Representations. p.~4. Toulon, France
  (2017), https://openreview.net/forum?id=H1ZaRZVKg

\bibitem{Winograd80}
Winograd, S.: Arithmetic Complexity Computations. {SIAM} Publications, Bristol,
  England (1980)

\bibitem{Zhao18}
Zhao, Y., Wang, D., Wang, L., Liu, P.: A faster algorithm for reducing the
  computational complexity of convolutional neural networks. Algorithms
  \textbf{11}(10) (2018). \doi{10.3390/a11100159},
  \url{http://www.mdpi.com/1999-4893/11/10/159}

\end{thebibliography}

\end{document}